\documentclass[10pt,twocolumn,letterpaper]{article}

\usepackage{cvpr}              

\usepackage{multirow}
\usepackage{mathrsfs}
\usepackage{xcolor}

\newcommand{\bftab}[1]{{\fontseries{b}\selectfont#1}}
\newcommand{\tabincell}[2]{\begin{tabular}{@{}#1@{}}#2\end{tabular}}
\makeatletter
\newcommand*\bigcdot{\mathpalette\bigcdot@{.5}}
\newcommand*\bigcdot@[2]{\mathbin{\vcenter{\hbox{\scalebox{#2}{$\m@th#1\bullet$}}}}}
\makeatother
\newcommand{\RED}[1]{\textcolor{BrickRed}{\bftab{#1}}}
\newcommand{\BLUE}[1]{\textcolor{NavyBlue}{\bftab{#1}}}
\newcommand{\GREEN}[1]{\textcolor{OliveGreen}{\bftab{#1}}}

\definecolor{cvprblue}{rgb}{0.21,0.49,0.74}
\usepackage[pagebackref,breaklinks,colorlinks,allcolors=cvprblue]{hyperref}

\def\paperID{*****} 
\def\confName{CVPR}
\def\confYear{2025}

\title{OpenAnimals: Revisiting Person Re-Identification for Animals\\Towards Better Generalization}

\author{Saihui Hou$^{1,3}$ \quad Panjian Huang$^1$ \quad Zengbin Wang$^2$ \quad Yuan Liu$^1$ \quad Zeyu Li$^1$ \\ Man Zhang$^2$\quad Yongzhen Huang$^{1,3}$\thanks{Corresponding author.}\\
$^1$School of Artificial Intelligence, Beijing Normal University\\
$^2$School of Artificial Intelligence, Beijing University of Posts and Telecommunications\\
$^3$WATRIX.AI
}

\begin{document}
\maketitle

\begin{abstract}
This paper addresses the challenge of animal re-identification, an emerging field that shares similarities with person re-identification but presents unique complexities due to the diverse species, environments and poses.
To facilitate research in this domain, we introduce OpenAnimals, a flexible and extensible codebase designed specifically for animal re-identification.
We conduct a comprehensive study by revisiting several state-of-the-art person re-identification methods, including BoT, AGW, SBS, and MGN, and evaluate their effectiveness on animal re-identification benchmarks such as HyenaID, LeopardID, SeaTurtleID, and WhaleSharkID.
Our findings reveal that while some techniques generalize well, many do not, underscoring the significant differences between the two tasks.
To bridge this gap, we propose ARBase, a strong \textbf{Base} model tailored for \textbf{A}nimal \textbf{R}e-identification, which incorporates insights from extensive experiments and introduces simple yet effective animal-oriented designs.
Experiments demonstrate that ARBase consistently outperforms existing baselines, achieving state-of-the-art performance across various benchmarks.
%
\end{abstract} 

\section{Introduction}
\label{sec:intro}

Analogous to the person re-identification task~\cite{ye2021deep}, computer vision-based animal re-identification aims to \emph{recognize individual animals within a specific species}.
Due to its non-invasive nature and the potential for automated recognition, animal re-identification holds significant promise in various wildlife research and conservation applications, such as population monitoring, behavioral studies, and the protection of endangered species~\cite{papafitsoros2021social,schofield2022more,vidal2021perspectives}.

\begin{figure}
  \centering
  \begin{subfigure}{0.45\linewidth}
    \includegraphics[width=1.0\linewidth]{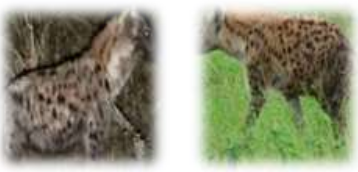}
    \caption{HyenaID}
    \label{HyenaID}
  \end{subfigure}
  \hfill
  \begin{subfigure}{0.45\linewidth}
    \includegraphics[width=1.0\linewidth]{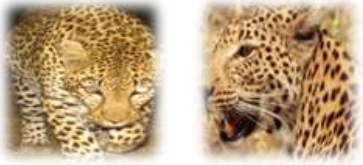}
    \caption{LeopardID}
    \label{LeopardID}
  \end{subfigure}
  \\
  \begin{subfigure}{0.45\linewidth}
    \includegraphics[width=1.0\linewidth]{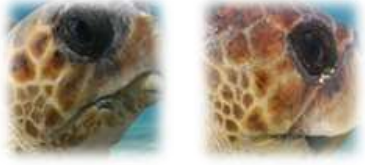}
    \caption{SeaTurtleID}
    \label{SeaTurtleID}
  \end{subfigure}
  \hfill
  \begin{subfigure}{0.45\linewidth}
    \includegraphics[width=1.0\linewidth]{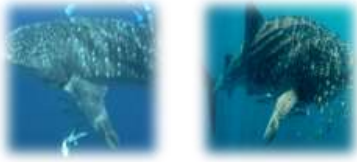}
    \caption{WhaleSharkID}
    \label{WhaleSharkID}
  \end{subfigure}
  \vspace{-1ex}
  \caption{Illustration of various species.}
  \label{fig:species}
  \vspace{-4ex}
\end{figure}

In the computer vision community, research on re-identification has predominantly focused on people~\cite{zheng2015scalable,wei2018person,ristani2016performance,zheng2017unlabeled} and vehicles~\cite{lou2019large,liu2016deep,liu2017provid}.
While person re-identification shares some conceptual similarities with animal re-identification, there are several notable differences between these two tasks, which include, but are not limited to, the following aspects:

\begin{itemize}
    \item \emph{Species Diversity}: Unlike person re-identification, which deals exclusively with human subjects, animal re-identification must accommodate a wide range of species, such as hyenas~\cite{HyenaID}, leopards~\cite{LeopardID}, sea turtles~\cite{SeaTurtuleID}, and whale sharks~\cite{WhaleSharkID}. Each species presents unique challenges in terms of visual appearance and behavior.

    \item \emph{Environmental Variability}: The living environments of different species vary dramatically, both spatially and temporally. For instance, some species inhabit grasslands~\cite{HyenaID,LeopardID}, while others are found in oceans~\cite{SeaTurtuleID,WhaleSharkID}. In contrast, datasets for person re-identification are predominantly collected in urban environments, where conditions are relatively controlled and consistent.

    \item \emph{Pose Variability}: The common poses exhibited by animals are highly species-specific. For example, quadrupedal locomotion in hyenas and leopards~\cite{HyenaID,LeopardID} is markedly different from the bipedal walking observed in humans, and swimming behaviors in marine species like sea turtles and whale sharks~\cite{SeaTurtuleID,WhaleSharkID} introduce further complexities.

    \item \emph{Data Availability}: Due to the inherent difficulties in data collection and annotation in wildlife settings, the amount of data available for animal re-identification is often limited~\cite{HyenaID,LeopardID,SeaTurtuleID,WhaleSharkID}. This scarcity of data poses significant challenges for training robust and generalized models.
\end{itemize}

Based on these observations, a natural and critical question arises: \textbf{\emph{Can the substantial research efforts and methodologies developed for person re-identification be effectively generalized to the domain of animal re-identification?}}

In this work, we commit ourselves to addressing the aforementioned question.
To begin with, we develop a flexible and extensible codebase for animal re-identification, named \textbf{OpenAnimals}, built on PyTorch.
This codebase is designed with two key principles in mind:
\begin{enumerate}[label=(\arabic*)]
\item \emph{Compatibility with Person Re-identification}: We ensure that OpenAnimals is fully compatible with state-of-the-art methods in person re-identification, facilitating the evaluation of the effectiveness of human-oriented designs when applied to animals.
\item \emph{Support for Multiple Species}: OpenAnimals is designed to support re-identification tasks across various animal species, such as the aforementioned hyenas~\cite{HyenaID}, leopards~\cite{LeopardID}, sea turtles~\cite{SeaTurtuleID}, and whale sharks~\cite{WhaleSharkID}, within a unified framework.
\end{enumerate}
Furthermore, leveraging OpenAnimals, we revisit several representative person re-identification methods, including BoT~\cite{luo2019bag}, AGW~\cite{ye2021deep}, SBS~\cite{he2023fastreid}, and MGN~\cite{wang2018learning}, and re-conduct ablation studies on these methods using animal re-identification benchmarks.
Interestingly, we observe phenomena that diverge from previous conclusions in the person re-identification domain, underscoring the significant differences between the two tasks.

Building on these observations and incorporating animal-specific insights, we propose a strong base model for animal re-identification, denoted as \textbf{ARBase}.
We systematically examine the entire re-identification pipeline, dividing it into modular stages, namely \emph{Data}, \emph{Backbone}, \emph{Head}, \emph{Loss}, and \emph{Training \& Testing}.
For each stage, we adopt conceptually simple yet empirically powerful designs to construct ARBase.
Without relying on complex modifications, ARBase achieves highly competitive performance across various animal benchmarks, providing a solid foundation for future research.
For instance, on the HyenaID~\cite{HyenaID} and WhaleSharkID~\cite{WhaleSharkID} benchmarks, ARBase outperforms the best person re-identification baselines by 14.54\% and 9.90\% in terms of rank-1 accuracy, respectively.

In summary, the main contributions of this work consists of three folds:
\begin{enumerate}[label=(\arabic*)]
\item We present an open-source codebase named OpenAnimals\footnote{The codebase is included as supplementary material for this submission. Please refer to the README file for installation and usage details.} for animal re-identification. This codebase is compatible with person re-identification methods and supports multiple animal species within a unified framework.
\item We revisit and analyze several representative person re-identification methods in the context of animal re-identification, leading to insightful observations.
\item We develop a strong base model, ARBase, for animal re-identification, which achieves promising performance across multiple benchmarks without relying on sophisticated designs.
%
\end{enumerate}


\section{Related Work}
\label{sec:related}

\paragraph{Person Re-Identification}
Person re-identification has been extensively studied as a specific retrieval problem across multiple camera views.
The objective is to accurately identify a person-of-interest either at different times within the same camera or across different cameras.
Based on the comprehensive survey by Ye \etal~\cite{ye2021deep}, the literature on person re-identification can be broadly categorized into four groups, incorporating recent advancements as follows.

\begin{itemize}
    \item \emph{Global Feature-Based Methods}: These approaches aim to extract a single global feature vector representing the entire person image.
    Early works in this category leverage deep learning models originally designed for image classification tasks~\cite{zheng2017person,wang2016joint,qian2017multi} and adapt them for person retrieval by training on large-scale person datasets.

    \item \emph{Local Feature-Based Methods}: Building upon global features, these methods focus on learning fine-grained, part-level features to capture more detailed and robust representations.
    The body parts are typically detected or segmented using techniques such as human parsing~\cite{guo2019beyond}, pose estimation~\cite{su2017pose}, or simple horizontal partitioning strategies~\cite{sun2018beyond}.

    \item \emph{Auxiliary Feature-Based Methods}: These approaches enhance global and local features by incorporating additional semantic information, such as camera viewpoints~\cite{liu2019view}, clothing attributes~\cite{su2016deep,lin2019improving}, and hairstyle features~\cite{tay2019aanet}.
    Moreover, this category encompasses methods that employ generative models to augment training data, improving model robustness and generalization~\cite{zheng2017unlabeled,liu2018pose,qian2018pose,zhong2018camera,zheng2019joint}.

    \item \emph{Video-Based Methods}: Extending beyond image-based analysis, these methods exploit the rich spatiotemporal information available in video sequences to enhance re-identification performance.
    Key challenges addressed include effectively capturing temporal dynamics~\cite{zang2022multidirection,gu2020appearance,hou2021bicnet,wang2021pyramid}, handling noisy or outlier frames~\cite{li2018diversity,subramaniam2019co}, and managing variable-length video sequences~\cite{fu2019sta,li2019global}.
\end{itemize}

In Section~\ref{sec:revisiting}, we delve into the details of several representative methods, including BoT~\cite{luo2019bag}, AGW~\cite{ye2021deep}, SBS~\cite{he2023fastreid}, and MGN~\cite{wang2018learning}, evaluating their applicability to animal re-identification tasks.
It is worth noting that recent works exploring text-driven~\cite{qin2024noisy,zuo2024ufinebench}, multimodal~\cite{guo2024lidar,ren2024implicit}, and lifelong learning~\cite{cui2024learning,xu2024distribution} approaches for person re-identification are beyond the scope of this study.
We acknowledge and appreciate the extensive efforts in prior research, which provide a strong foundation for advancing animal re-identification methodologies.

\paragraph{Animal Re-Identification}
Despite its critical importance for biological and ecological studies, research in animal re-identification has lagged behind that of person re-identification.
Our work is motivated by the need to bridge this gap and advance the state-of-the-art in this domain.
Existing approaches for animal re-identification can be roughly categorized into two main groups:

\begin{itemize}
    \item \emph{Hand-Crafted Feature-Based Methods}: Traditional visual descriptors, such as SIFT~\cite{lowe2004distinctive} and SURF~\cite{bay2006surf}, have been widely employed in early animal re-identification systems like WildID~\cite{bolger2012computer} and HotSpotter~\cite{crall2013hotspotter}.
    While effective to some extent, these hand-crafted features exhibit limitations in achieving high performance and scaling to large, diverse datasets~\cite{krizhevsky2012imagenet,he2016deep}.

    \item \emph{Deep Learning-Based Methods}: More recent approaches leverage deep learning models for feature extraction and representation learning in animal re-identification tasks.
    Many studies adapt existing architectures from face recognition or person re-identification with minimal modifications.
    For instance, standard backbone networks trained with triplet loss~\cite{schroff2015facenet,hermans2017defense} or ArcFace loss~\cite{deng2019arcface} are utilized in~\cite{vcermak2024wildlifedatasets} to identify individuals across various species.
    Additionally, part-based models guided by pose information have been explored for specific species recognition, such as tigers in~\cite{ATRW}.
    Some works also employ pre-trained foundational models, like CLIP~\cite{radford2021learning} and DINOv2~\cite{oquab2023dinov2}, to extract robust features for animal re-identification.
\end{itemize}

While these efforts have provided valuable initial insights, there remains a need for more comprehensive analyses and the development of task-specific designs tailored to the unique challenges of animal re-identification.
Our study seeks to address these gaps by systematically evaluating existing methods and proposing effective solutions grounded in animal-specific contexts.


\section{Our Approach}
Through a comprehensive comparison with person re-identification, we identify three key challenges that must be urgently addressed for effective animal re-identification:

\begin{enumerate}[label=(\arabic*)]
    \item \textbf{Lack of a Flexible Codebase:}
    There is a critical need to develop an animal-oriented codebase for re-identification that can integrate specialized algorithms and facilitate rigorous ablation studies. Such a codebase would serve as a foundational tool for researchers in this field.

    \item \textbf{Unclear Generalization from Person Re-ID:}
    It remains unclear which techniques and methodologies that have been successful in person re-identification can be effectively generalized to animal re-identification. This lack of clarity hinders the application of existing methods to new domains.

    \item \textbf{Necessity of a Strong Base Model:}
    There is an urgent need to design a robust baseline for animal re-identification across various species. Such a model would provide a solid starting point for advanced research and development in this area.
\end{enumerate}

Our work is driven by the motivation to address these challenges. In the following sections, we will first introduce the construction of \textbf{OpenAnimals}, a flexible and extensible codebase designed specifically for animal re-identification.
Next, we will revisit and evaluate the key designs of several representative person re-identification methods within the context of animal re-identification.
Finally, we will present \textbf{ARBase}, a strong base model that has been developed to meet the unique challenges of animal re-identification and to serve as a reliable foundation for future research.

\subsection{OpenAnimals}
In this section, we elaborate on the design principles and implementation details of \textbf{OpenAnimals}, a flexible and extensible platform for animal re-identification.
As previously mentioned, OpenAnimals is developed based on two core principles:

\begin{enumerate}[label=(\arabic*)]
    \item \emph{Compatibility with Person Re-Identification:}
    In recent years, person re-identification has garnered significant attention in the computer vision community, leading to the development of numerous advanced methodologies.
    Despite the differences between person and animal re-identification, as analyzed earlier, both tasks share fundamental similarities in their objectives and problem definitions.
    Therefore, it is both natural and critical to leverage the advancements made in person re-identification to recognize individual animals within each species.

    \item \emph{Support for Multiple Species:}
    A key distinction in animal re-identification compared to person re-identification is the necessity to address the diversity of species, each with unique living environments and common poses.
    To date, there are over 30 benchmarks covering different species that have been published and are available for re-identification tasks~\cite{vcermak2024wildlifedatasets}.
    We aim for OpenAnimals to support re-identification across all these species using a unified framework.
\end{enumerate}

\begin{figure}[t]
	\centering
	\includegraphics[width=0.90\linewidth]{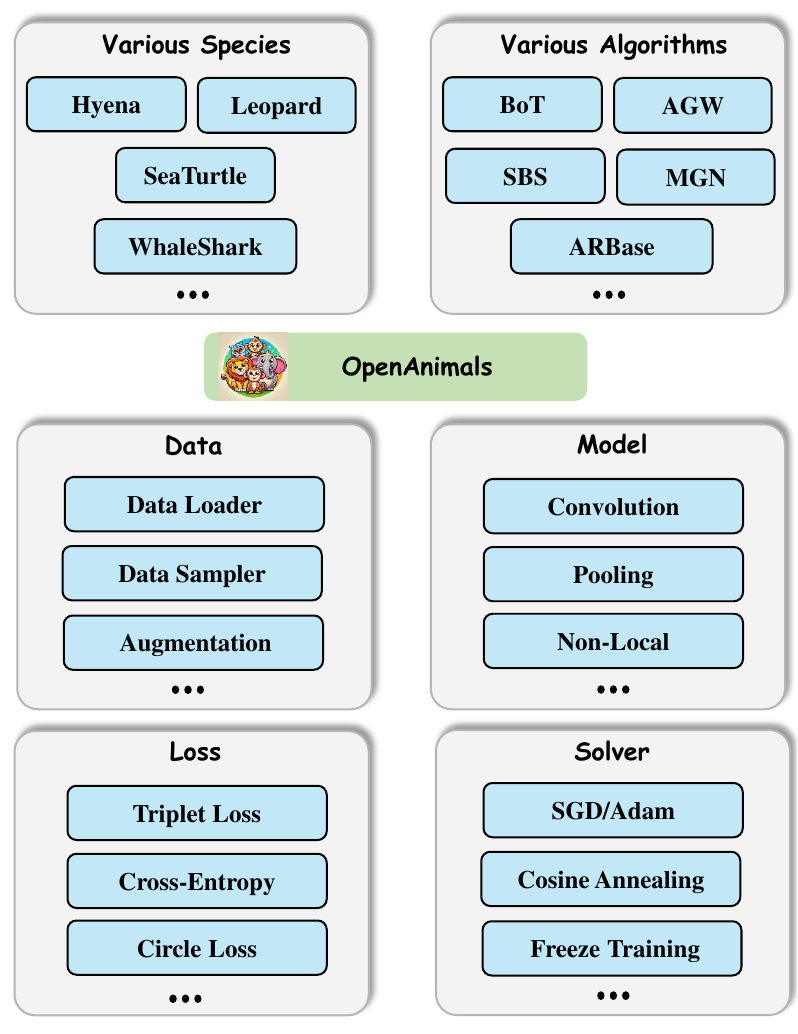}
	\caption{
    Illustration of main modules in OpenAnimals.
	}
	\label{fig:openanimals_modules}
\end{figure}

To fulfill these requirements, we conducted a comprehensive literature review and constructed OpenAnimals by leveraging the recent advancements of FastReID\footnote{\url{https://github.com/JDAI-CV/fast-reid}}~\cite{he2023fastreid} and WildLifeDatasets\footnote{\url{https://github.com/WildlifeDatasets/wildlife-datasets}}~\cite{vcermak2024wildlifedatasets}.
First, OpenAnimals inherits the core layers from FastReID to ensure compatibility with person re-identification, with essential modifications made to tailor it for animal re-identification.
Second, OpenAnimals achieves support for multiple species by incorporating the dataset organization strategies from WildLifeDatasets.
We acknowledge that the development of OpenAnimals has been significantly facilitated by the excellent work done in FastReID and WildLifeDatasets, and we deeply appreciate their contributions.

We envision OpenAnimals as a valuable tool for advancing research in animal re-identification, with future plans to integrate more animal-oriented designs into the platform.
Figure~\ref{fig:openanimals_modules} illustrates the main modules of OpenAnimals, which align with the design styles of most PyTorch-based codebases~\cite{he2023fastreid,fan2023opengait} to ensure high usability and feasibility.

\begin{table*}[!tbp]
\setlength\tabcolsep{12pt} 
	\caption{
	Revisiting core designs in BoT~\cite{luo2019bag} and AGW~\cite{ye2021deep} of person re-identification for animals.
    The \RED{red} (\GREEN{green}) results are \RED{lower} (\GREEN{higher}) than the references.
    \emph{R1} for \emph{Rank-1 Accuracy}, \emph{mAP} for \emph{mean Average Precision}.
	}
	\label{tab:revisit_BoT_AGW}
	\begin{center}
		\resizebox{0.85\linewidth}{!}{%
			\begin{tabular}{c|cc|cc|cc|cc}
    		\toprule
    		\multirow{2}{*}{Method} & \multicolumn{2}{c|}{HyenaID~\cite{HyenaID}} & \multicolumn{2}{c|}{LeopardID~\cite{LeopardID}} & \multicolumn{2}{c|}{SeaTurtleID~\cite{SeaTurtuleID}} & \multicolumn{2}{c}{WhaleSharkID~\cite{WhaleSharkID}} \\
    		& R1 & mAP & R1 & mAP & R1 & mAP & R1 & mAP \\
    		\midrule
BoT~\cite{luo2019bag} & 58.64 & 34.96 & 54.92 & 27.65 & 84.01 & 41.92 & 52.54 & 20.86 \\
\midrule
\emph{w/o Random Erasing} & \GREEN{58.64} & \GREEN{35.01} & \GREEN{54.92} & \GREEN{28.31} & \GREEN{84.30} & \GREEN{45.32} & \RED{47.46} & \RED{19.68} \\
\emph{w/o Label Smoothing} & \RED{52.73} & \RED{31.03} & \GREEN{56.56} & \GREEN{29.15} & \RED{79.36} & \RED{39.40} & \RED{50.76} & \RED{19.70} \\
\emph{w/o Last Stride} & \RED{52.27} & \RED{31.68} & \RED{53.28} & \RED{27.11} & \RED{78.20} & \RED{38.85} & \RED{47.72} & \RED{18.79} \\
\emph{w/o BNNeck} & \RED{54.55} & \RED{32.18} & \RED{52.87} & \RED{26.63} & \RED{77.03} & \RED{36.22} & \RED{45.18} & \RED{18.33} \\
    		\midrule
AGW~\cite{ye2021deep} & 56.36 & 32.72 & 54.10 & 28.67 & 85.17 & 46.18 & 50.76 & 21.11 \\
\midrule
\emph{w/o Non-local Attention} & \RED{55.91} & \GREEN{33.04} & \RED{51.64} & \RED{27.39} & \RED{83.72} & \RED{43.98} & \GREEN{52.28} & \RED{20.34} \\
\emph{w/o Generalized-mean Pooling} & \RED{55.45} & \GREEN{33.14} & \GREEN{56.15} & \RED{27.55} & \RED{84.88} & \RED{44.27} & \RED{49.75} & \RED{19.75} \\
\emph{w/o Weighted Triplet Loss} & \RED{55.45} & \GREEN{33.80} & \GREEN{54.51} & \RED{26.65} & \RED{83.43} & \RED{43.70} & \GREEN{52.03} & \GREEN{21.20} \\
    		\bottomrule
			\end{tabular}
		}
	\end{center}
\end{table*}

\subsection{Revisiting Person Re-ID for Animals}
\label{sec:revisiting}
The development of OpenAnimals is a crucial prerequisite for conducting revisiting analyses of person re-identification methods in the context of animal re-identification.
In this section, we leverage OpenAnimals to reassess the applicability of person re-identification techniques to animal re-identification, using representative methods such as BoT~\cite{luo2019bag}, AGW~\cite{ye2021deep}, SBS~\cite{he2023fastreid}, and MGN~\cite{wang2018learning}.
For this evaluation, we select HyenaID~\cite{HyenaID}, LeopardID~\cite{LeopardID}, SeaTurtleID~\cite{SeaTurtuleID}, and WhaleSharkID~\cite{WhaleSharkID} as benchmarks based on the following considerations:
(1) The first two species are terrestrial, while the latter two are aquatic, allowing us to explore the method's performance across different environments.
(2) These benchmarks are collected in the wild, introducing more challenging covariates such as varying lighting conditions, occlusions, and background clutter.
(3) Each benchmark contains a relatively higher number of samples averaged for each animal identity (\ie, $>10$), providing a robust basis for evaluation.

Unless otherwise specified, we adopt the following experimental settings:
(1) We use a disjoint-set partition, meaning that the identities in the training and test sets do not overlap.
%
(2) ResNet50 is used as the default backbone for all models.
(3) For evaluation, we select two samples from each identity as the query set and treat the remaining samples as the gallery.
(4) The evaluation metrics include rank-1 accuracy (R1) and mean Average Precision (mAP), which are widely used in person re-identification~\cite{ye2021deep}.
(5) We adhere to the configuration settings for each method as implemented in FastReID and use the reproduced performance as a reference.
(6) In the revisiting experiments, results that fall below the reference performance are highlighted in \RED{red}, while those that exceed the reference are highlighted in \GREEN{green}.

\subsubsection{Revisiting BoT~\cite{luo2019bag}}
BoT~\cite{luo2019bag} is well-regarded for its combination of effective strategies in person re-identification.
Despite its simplicity, the tricks introduced in BoT have been widely adopted in subsequent research.
Specifically, the key strategies investigated in BoT~\cite{luo2019bag} include:
\begin{enumerate}[label=(\arabic*)]
    \item \emph{Random Erasing}: A data augmentation technique that randomly erases a rectangular region in the input image, addressing occlusion issues and enhancing the model's generalization capability.
    \item \emph{Label Smoothing}: This technique transforms one-hot labels into a soft version, reducing the risk of overfitting by preventing the model from becoming too confident in its predictions.
    \item \emph{Last Stride}: Modifying the last stride of the backbone from $2$ to $1$ increases the spatial resolution of the feature maps generated by the backbone, allowing for finer-grained feature extraction.
    \item \emph{BNNeck}: This involves using a Batch Normalization (BN) layer to separate features used for triplet loss and cross-entropy loss.
    %
\end{enumerate}

In the first part of Table~\ref{tab:revisit_BoT_AGW}, we revisit these tricks on four animal re-identification benchmarks, and get the following conclusions:
\begin{enumerate}[label=(\alph*)]
\item Somewhat surprisingly, \emph{Random Erasing} instead has negative effect on the re-identification on HyenaID, LeopardID, SeaTurtleID.
A possible reason is that the earsing is likely to disrupt the critical details for animal re-identification since the differences between animal individuals are much more subtle as illustrated in Figure~\ref{fig:species}.

\item \emph{Label Smooth}, \emph{Last Stride}, and \emph{BNNeck} are beneficial on at least three datasets, which are basically consistent with the observations on person re-identification.
\end{enumerate}

\subsubsection{Revisiting AGW~\cite{ye2021deep}}
AGW~\cite{ye2021deep} is another strong baseline for person re-identification, building upon the foundations established by BoT~\cite{luo2019bag}.
AGW introduces three major improvements:
\begin{enumerate}[label=(\arabic*)]
    \item \emph{Non-Local Attention}: This block enhances the backbone by computing a weighted sum of features across all positions through self-attention, allowing the model to capture long-range dependencies and contextual information.
    \item \emph{Generalized-mean Pooling}: It represents a learnable pooling layer that replaces the commonly used max pooling or average pooling. It introduces an additional hyper-parameter that is optimized during the back-propagation process, enabling more flexible feature aggregation.
    \item \emph{Weighted Triplet Loss}: As a variant of the traditional triplet loss~\cite{hermans2017defense}, this method incorporates a weighting strategy for different triplets. It retains the advantage of optimizing relative distances between positive and negative pairs without requiring additional margin parameters.
\end{enumerate}

In the second part of Table~\ref{tab:revisit_BoT_AGW}, we evaluate these components on animal re-identification benchmarks.
Unfortunately, our experiments reveal that these components do not generalize well across different species, with none showing consistent effectiveness on at least three of the benchmarks.
This inconsistency underscores the significant differences between person and animal re-identification, highlighting the need for tailored approaches in animal re-identification.

\subsubsection{Revisiting SBS~\cite{he2023fastreid}}
SBS~\cite{he2023fastreid} is an unpublished method available in the FastReID repository.
It introduces the following key designs on top of BoT~\cite{luo2019bag}:
\begin{enumerate}[label=(\arabic*)]
    \item \emph{Freeze Training}: This technique locks the backbone parameters for a few iterations at the beginning of the training phase, potentially stabilizing the initial learning process.
    \item \emph{AutoAug}: A data augmentation strategy that automatically searches for improved augmentation policies within a predefined search space, aiming to enhance model generalization.
    \item \emph{Cosine Annealing}: This learning rate schedule adjusts the learning rate for each parameter group using a cosine annealing strategy, which gradually reduces the learning rate over time.
    \item \emph{Soft Margin Triplet Loss}: A variant of the triplet loss that replaces the fixed margin hyper-parameter with a soft margin, potentially allowing for more flexible optimization.
\end{enumerate}

We revisit these designs using a similar experimental setup. The results are summarized in the first part of Table~\ref{tab:revisit_SBS_MGN}.
From our experiments, we can draw the following conclusions:
\begin{enumerate}[label=(\alph*)]
    \item Contrary to person re-identification, the removal of \emph{Freeze Training}, \emph{AutoAug}, or \emph{Soft Margin Triplet Loss} actually leads to performance improvements on most animal re-identification datasets.
    \item \emph{Cosine Annealing} consistently proves to be beneficial across most experiments, suggesting its general utility for animal re-identification tasks.
\end{enumerate}

\begin{table*}[!tbp]
\setlength\tabcolsep{12pt} 
	\caption{
	Revisiting core designs in SBS~\cite{he2023fastreid} and MGN~\cite{wang2018learning} of person re-identification for animals.
    The \RED{red} (\GREEN{green}) results are \RED{lower} (\GREEN{higher}) than the references.
    \emph{R1} for \emph{Rank-1 Accuracy}, \emph{mAP} for \emph{mean Average Precision}.
	}
	\label{tab:revisit_SBS_MGN}
	\begin{center}
		\resizebox{0.85\linewidth}{!}{%
			\begin{tabular}{c|cc|cc|cc|cc}
    		\toprule
    		\multirow{2}{*}{Method} & \multicolumn{2}{c|}{HyenaID~\cite{HyenaID}} & \multicolumn{2}{c|}{LeopardID~\cite{LeopardID}} & \multicolumn{2}{c|}{SeaTurtleID~\cite{SeaTurtuleID}} & \multicolumn{2}{c}{WhaleSharkID~\cite{WhaleSharkID}} \\
    		& R1 & mAP & R1 & mAP & R1 & mAP & R1 & mAP \\
    		\midrule
SBS~\cite{he2023fastreid} & 51.82 & 30.56 & 51.23 & 26.54 & 84.01 & 44.63 & 47.46 & 18.84 \\
\midrule
\emph{w/o Freeze Training} & \GREEN{56.82} & \GREEN{30.80} & \GREEN{53.69} & \GREEN{27.58} & \RED{83.43} & \GREEN{45.44} & \RED{44.42} & \GREEN{19.09} \\
\emph{w/o AutoAug} & \RED{50.91} & \RED{29.80} & \GREEN{52.46} & \GREEN{27.94} & \GREEN{84.30} & \GREEN{44.97} & \GREEN{47.72} & \GREEN{19.14} \\
\emph{w/o Cosine Annealing} & \GREEN{51.82} & \GREEN{30.56} & \RED{48.36} & \RED{23.93} & \RED{82.27} & \RED{41.86} & \RED{46.19} & \RED{18.03} \\
\emph{w/o Soft Margin Triplet Loss} & \GREEN{52.27} & \RED{30.15} & \GREEN{52.87} & \GREEN{28.06} & \GREEN{84.01} & \RED{44.08} & \RED{45.94} & \GREEN{18.88} \\
    		\midrule
MGN~\cite{wang2018learning} & 55.91 & 31.08 & 53.69 & 28.21 & 86.05 & 46.67 & 50.25 & 21.47 \\
\midrule
\emph{w/o 3Parts Branch} & \RED{52.27} & \RED{30.55} & \RED{52.87} & \RED{27.12} & \RED{84.01} & \RED{45.52} & \RED{49.24} & \RED{20.47} \\
\emph{w/o 2Parts Branch} & \GREEN{56.82} & \GREEN{31.27} & \RED{51.23} & \RED{27.45} & \RED{84.30} & \RED{45.13} & \RED{49.24} & \RED{20.03} \\
\emph{w/o 3Parts/2Parts Branch} & \RED{48.18} & \RED{27.06} & \RED{48.77} & \RED{25.00} & \RED{81.98} & \RED{41.91} & \RED{46.19} & \RED{18.08} \\
    		\bottomrule
			\end{tabular}
		}
	\end{center}
\end{table*}

\subsubsection{Revisiting MGN~\cite{wang2018learning}}
MGN~\cite{wang2018learning} is a notable method that combines global and part-based features by horizontally splitting the feature maps output by the backbone network.
The core innovation lies in its multi-branch architecture, which enables the simultaneous extraction of both coarse-grained and fine-grained features.
In addition to the global branch, MGN includes two additional branches: the \emph{2Parts Branch}, which extracts 2-part features, and the \emph{3Parts Branch}, which extracts 3-part features, both aimed at improving person retrieval performance.

In the second part of Table~\ref{tab:revisit_SBS_MGN}, we extend this approach to animal re-identification by revisiting the effectiveness of the multi-branch architecture.
Interestingly, the results indicate that multi-granularity feature extraction remains beneficial for animals, leading to improved recognition performance on at least three of the benchmarks.

\subsubsection{Summary}
Through extensive revisiting experiments, we observe that many techniques developed for person re-identification exhibit limited generalization when directly applied to animal re-identification.
This discrepancy can be attributed to the significant differences between the two tasks, as discussed earlier.
However, we also identify several useful techniques that generalize well across different species.
The challenge now lies in effectively combining these insights to design an animal-oriented model for re-identification, which is an urgent need in advancing this field.

\begin{figure*}[t]
	\centering
	\includegraphics[width=0.99\linewidth]{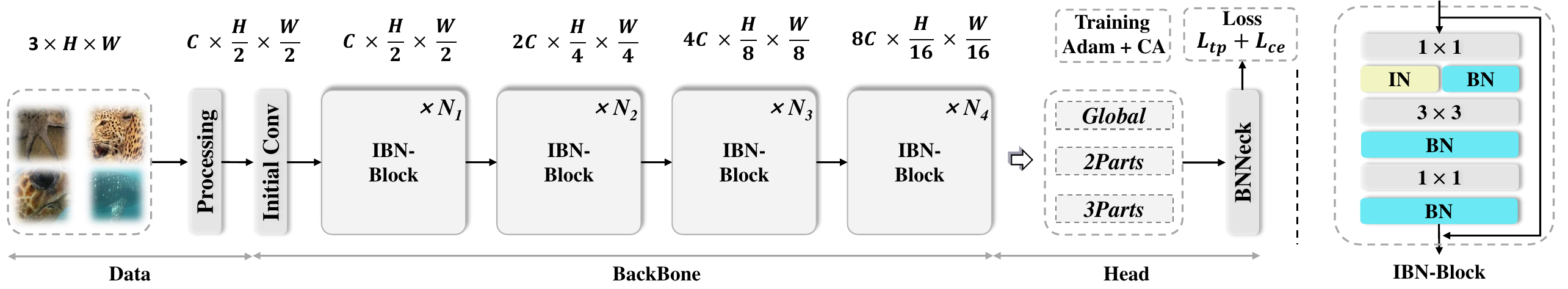}
	\caption{
   Illustration of ARBase.
   \emph{IBN}, \emph{IN} and \emph{BN} for \emph{Instance-Batch}, \emph{Instance} and \emph{Batch Normalization},
   $[N_{1}, N_{2}, N_{3}, N_{4}]$ for number of blocks in each stage (\eg, $[3, 4, 6, 3]$ for ResNet50),
   \emph{CA} for \emph{Cosine Annealing},
   $L_{tp}$ and $L_{ce}$ for \emph{Triplet} and \emph{Cross-Entropy Loss}.
	}
	\label{fig:ARBase}
\end{figure*}

\subsection{ARBase: A Strong Baseline for Animal Re-ID}
In this section, we aim to establish a strong baseline, referred to as ARBase, for animal re-identification.
To facilitate modular design and ensure flexibility, we divide the entire pipeline into five main modules, each of which will be elaborated on in the following sections.
Our primary goal is to avoid introducing overly complex designs, thereby ensuring that ARBase generalizes well across various species.
The implementation of ARBase will be released alongside OpenAnimals.

\subsubsection{Data}
The \emph{Data} module is responsible for sampling datasets and performing data augmentation.
Here, we highlight key differences from person re-identification:

\begin{enumerate}[label=(\arabic*)]
    \item An often overlooked but critical insight relates to input resolution.
    Since most human subjects are captured while standing upright, almost all person re-identification methods use input resolutions where the \emph{width is less than the height} (\eg, $[256, 128]$ for BoT~\cite{luo2019bag}, $[384, 128]$ for MGN~\cite{wang2018learning}).
    However, this prior knowledge does not apply to animals, as each species exhibits unique common poses, as illustrated in Figure~\ref{fig:species}.
    Therefore, we highly recommend using resolutions where \emph{the width is equal to or greater than the height} (\eg, $[384, 384]$ in our experiments).
    This simple modification has a significant impact on recognition performance across various species.

    \item Based on our experimental comparisons revisiting BoT~\cite{luo2019bag} and SBS~\cite{he2023fastreid}, we choose to omit \emph{Random Erasing} and \emph{AutoAug} for data augmentation\footnote{While it is technically feasible to fine-tune the hyper-parameters of \emph{Random Erasing} and the search space of \emph{AutoAug} for animals, we omit them for simplicity.} and instead only apply random horizontal flips with a probability of $0.5$.
\end{enumerate}

\subsubsection{Backbone}
The \emph{Backbone} module extracts identity-related features from the input data.
Considering the balance between effectiveness and efficiency, we use ResNet-50 pretrained on ImageNet as the backbone and introduce the following enhancements:

\begin{enumerate}[label=(\arabic*)]
    \item We modify the last stride to $1$, following BoT~\cite{luo2019bag} (Table~\ref{tab:revisit_BoT_AGW}), to obtain higher resolution feature maps that capture fine-grained identity-related details.
    \item We replace Batch Normalization (BN) with Instance-Batch Normalization (IBN)~\cite{pan2018two}, which combines Instance Normalization (IN) and Batch Normalization (BN).
    The rationale is that IN helps learn features invariant to appearance changes (\eg, different environments), while BN is essential for preserving content-related information.
    \item Inspired by the observations from revisiting MGN~\cite{wang2018learning}, we introduce a multi-branch architecture to simultaneously learn coarse-grained (\emph{global}) and fine-grained (\emph{2-part} and \emph{3-part}) features.
\end{enumerate}

\subsubsection{Head}
The \emph{Head} module processes the features extracted by the \emph{Backbone} and prepares them for loss computation.
To maintain simplicity and generalization, the \emph{Head} primarily consists of a \emph{Global Average Pooling} layer for spatial feature aggregation and a \emph{Linear} layer for further mapping and channel transformation.
Additionally, we employ the BNNeck~\cite{luo2019bag} to separate the final output into two feature spaces, facilitating the computation of triplet loss and cross-entropy loss independently.

\begin{table*}[!tbp]
\setlength\tabcolsep{8pt} 
	\caption{
	Performance comparison.
    In each column, the best result is marked in \textbf{bold} and the superscript indicates the improvement compared to the second best marked with \underline{underline}.
    \emph{R1} for \emph{Rank-1 Accuracy}, \emph{mAP} for \emph{mean Average Precision}.
	}
	\label{tab:performance_comparison}
	\begin{center}
		\resizebox{0.90\linewidth}{!}{%
			\begin{tabular}{c|cc|cc|cc|cc}
    		\toprule
    		\multirow{2}{*}{Method} & \multicolumn{2}{c|}{HyenaID~\cite{HyenaID}} & \multicolumn{2}{c|}{LeopardID~\cite{LeopardID}} & \multicolumn{2}{c|}{SeaTurtleID~\cite{SeaTurtuleID}} & \multicolumn{2}{c}{WhaleSharkID~\cite{WhaleSharkID}} \\
    		& R1 & mAP & R1 & mAP & R1 & mAP & R1 & mAP \\
    		\midrule
BoT~\cite{luo2019bag} & \underline{58.64} & \underline{34.96} & \underline{54.92} & 27.65 & 84.01 & 41.92 & \underline{52.54} & 20.86 \\
AGW~\cite{ye2021deep} & 56.36 & 32.72 & 54.10 & \underline{28.67} & 85.17 & 46.18 & 50.76 & 21.11 \\
SBS~\cite{he2023fastreid} & 51.82 & 30.56 & 51.23 & 26.54 & 84.01 & 44.63 & 47.46 & 18.84 \\
MGN~\cite{wang2018learning} & 55.91 & 31.08 & 53.69 & 28.21 & \underline{86.05} & \underline{46.67} & 50.25 & \underline{21.47} \\
ARBase(\bftab{Ours}) & \bftab{73.18}$^{\BLUE{\uparrow 14.54}}$ & \bftab{44.87}$^{\BLUE{\uparrow 9.91}}$ & \bftab{64.34}$^{\BLUE{\uparrow 9.42}}$ & \bftab{37.08}$^{\BLUE{\uparrow 8.41}}$ & \bftab{86.92}$^{\BLUE{\uparrow 0.87}}$ & \bftab{55.99}$^{\BLUE{\uparrow 9.32}}$ & \bftab{62.44}$^{\BLUE{\uparrow 9.90}}$ & \bftab{29.45}$^{\BLUE{\uparrow 7.98}}$ \\
    		\bottomrule	
			\end{tabular}
		}
	\end{center}
\end{table*}

\subsubsection{Loss}
The \emph{Loss} module supervises the entire training process.
We adopt a combination of triplet loss $L_{tp}$ and cross-entropy loss $L_{ce}$, both of which are widely used in person re-identification.
Specifically, the features before BNNeck are used to compute $L_{tp}$, while those after BNNeck are used to compute class scores for $L_{ce}$:
\begin{equation}
\label{eq_triplet_ce}
\begin{aligned}
\small
L_{{tp}} =&  \frac{1}{ N_{b} } \sum_{i=1}^{N_{b}} \mathrm{max}(0, m + d_{pos}^{i} - d_{neg}^{i}) \\
L_{{ce}} =&  - \frac{1}{ N_{b} } \sum_{i=1}^{N_{b}} \sum_{c=1}^{N_{c}} q_{c}^{i} \log p_{c}^{i} \\
\end{aligned}
\end{equation}
where $N_{b}$ is the number of training samples in a batch and $N_{c}$ is the number of identities in the training set, with each identity regarded as a class.
For triplet loss, $m$ is a margin hyper-parameter, and $d_{pos}^{i}$ and $d_{neg}^{i}$ represent the largest positive pair distance and the smallest negative pair distance for a sample~\cite{hermans2017defense} (\ie, hard example mining within the triplets).
For cross-entropy loss, $p_{c}^{i}$ denotes the predicted probabilities for the training identities, and $q_{c}^{i}$ is the soft version of one-hot identity labels generated by label smoothing~\cite{luo2019bag}:
\begin{equation}
\label{eq_label_smooth}
\small
	q_{c}^{i} =\left\{
	\begin{aligned}
		& 1 - \epsilon~~~~~~~~~~\mathrm{if}~c=y\\
		& \frac{\epsilon}{N_{c}-1}~~~~~~\mathrm{otherwise}
	\end{aligned}
	\right.
\end{equation}
where $y$ is the index of the ground-truth class, and $\epsilon$ is a small constant that encourages the model to be less confident during training.

\subsubsection{\emph{Training \& Testing}}
\emph{Training \& Testing} encompasses the optimization strategies used during training and the procedures for making predictions on unseen samples.
For training, we use the popular Adam optimizer, as in BoT~\cite{luo2019bag}, and adopt a cosine annealing strategy to adjust the learning rate, following the approach in SBS~\cite{he2023fastreid}.
For testing, we use the features before BNNeck as embeddings for each sample to match the probe and gallery sets.
Each probe sample is assigned the identity label of the gallery sample with the smallest Euclidean distance.
As in the revisiting experiments, rank-1 accuracy (R1) and mean Average Precision (mAP) are used as evaluation metrics.

\paragraph{Discussion}
%
%
ARBase distinguishes itself through animal-oriented designs derived from extensive experiments on person re-identification and insightful task-specific observations.
Our approach is similar in spirit to MGN~\cite{wang2018learning} for person re-identification and GaitBase~\cite{fan2023opengait} for gait recognition.
Without relying on complex modifications, these methods achieve highly promising performance compared to state-of-the-art counterparts and are widely adopted in practical systems due to their conceptually simple designs and strong generalization abilities.
We hope that ARBase will serve as a solid starting point and save valuable effort for future advancements in animal re-identification. 

\section{Experiments}
\label{sec:experiments}
\subsection{Setup}
All models are implemented using the OpenAnimals framework, and we apply a unified setting across all animal datasets.
The details about the datasets and implementation are provided in the supplementary material.

\subsection{Performance Comparison}
In Table~\ref{tab:performance_comparison}, we compare the performance of ARBase on four animal benchmarks with several person re-identification baselines, including BoT~\cite{luo2019bag}, AGW~\cite{ye2021deep}, SBS~\cite{he2023fastreid}, and MGN~\cite{wang2018learning}.
From the results, we can draw the following observations:
\begin{enumerate}[label=(\arabic*)]
    \item Despite achieving remarkable accuracy in person re-identification, these baselines do not generalize well to animal benchmarks, highlighting the significant differences between the two tasks.
    Among the four baselines, the simple yet effective BoT method achieves the best rank-1 accuracy on HyenaID~\cite{HyenaID}, LeopardID~\cite{LeopardID}, and WhaleSharkID~\cite{WhaleSharkID}, which further validates the necessity of our revisiting experiments to reassess the applicability of person re-identification techniques to animal re-identification.
    \item Notably, ARBase achieves state-of-the-art performance without relying on complex modifications and outperforms the baselines by a significant margin.
    For example, compared to the second-best results, ARBase improves rank-1 accuracy by 14.54\% on HyenaID~\cite{HyenaID}, 9.42\% on LeopardID~\cite{LeopardID}, and 9.90\% on WhaleSharkID~\cite{WhaleSharkID}, demonstrating its effectiveness and robustness as a strong baseline.
\end{enumerate}

\begin{table}[!tbp]
\setlength\tabcolsep{4pt} 
	\caption{
	Ablation study on \emph{Data} and \emph{Backbone}.
    \emph{IBN} for \emph{Instance-Batch Normalization}, \emph{MB} for \emph{Multi-Branch Architecture}.
	}
	\label{tab:ablation_data_backbone}
\vspace{-3ex}
	\begin{center}
		\resizebox{0.90\linewidth}{!}{%
			\begin{tabular}{c|c|cc|cc}
    		\toprule
    		\multirow{2}{*}{Method} & \multirow{2}{*}{\tabincell{c}{Input\\Resolution}} & \multicolumn{2}{c|}{HyenaID} & \multicolumn{2}{c}{WhaleSharkID} \\
    		& & R1 & mAP & R1 & mAP \\
    		\midrule
    		BoT~\cite{luo2019bag} 		 &  [256,128] & 58.64 & 34.96 & 52.54 & 20.86 \\
    		AGW~\cite{ye2021deep} 		 &  [256,128] & 56.36 & 32.72 & 50.76 & 21.11 \\		
    		SBS~\cite{he2023fastreid} 	 &  [384,128] & 51.82 & 30.56 & 47.46 & 18.84 \\
    		MGN~\cite{wang2018learning}  &  [384,128] & 55.91 & 31.08 & 50.25 & 21.47 \\
            \midrule
    		BoT~\cite{luo2019bag} 		 &  [384,384] & 60.45 & 36.43 & 58.12 & 24.39 \\
    		AGW~\cite{ye2021deep} 		 &  [384,384] & 62.73 & 37.36 & 58.38 & 26.21 \\		
    		SBS~\cite{he2023fastreid} 	 &  [384,384] & 57.27 & 33.01 & 56.60 & 24.71 \\
    		MGN~\cite{wang2018learning}  &  [384,384] & 61.36 & 35.81 & 58.63 & 26.19 \\
            \midrule
            ARBase(\emph{w/o IBN})       & [384,384] & 69.09 & 43.58 & 61.93 & 29.28 \\
            ARBase(\emph{w/o MB})        & [384,384] & 71.36 & 42.87 & 61.42 & 27.78 \\
    		ARBase(\bftab{Ours})         & [384,384] & \bftab{73.18} & \bftab{44.87} & \bftab{62.44} & \bftab{29.45} \\
    		\bottomrule	
			\end{tabular}
		}
	\end{center}
\vspace{-4ex}
\end{table}

\subsection{Ablation Study}
To evaluate the effectiveness of the animal-oriented designs in the five modules of ARBase, we conduct a comprehensive ablation study across all the mentioned datasets\footnote{Due to space constraints, the ablation studies on LeopardID~\cite{LeopardID} and SeaTurtleID~\cite{SeaTurtuleID} are provided in the supplementary material.}.

\paragraph{Ablation Study on \emph{Data} and \emph{Backbone}.}
Table~\ref{tab:ablation_data_backbone} presents the ablation studies on the \emph{Data} and \emph{Backbone} modules.
Specifically, our insights into adjusting the input aspect ratio to accommodate diverse animal poses can also be applied to the baseline models, leading to notable performance improvements.
As shown in Table~\ref{tab:ablation_data_backbone}, these enhancements significantly boost the performance of the baselines, although they still fall short of ARBase.
Notably, the last three rows of Table~\ref{tab:ablation_data_backbone} demonstrate the impact of integrating IBN into the backbone and the use of a multi-branch architecture for extracting both coarse-grained and fine-grained features, further contributing to the effectiveness of the re-identification process.

\paragraph{Ablation Study on \emph{Head}, \emph{Loss}, and \emph{Training \& Testing}.}
The ablation studies on the \emph{Head}, \emph{Loss}, and \emph{Training \& Testing} modules are presented in Table~\ref{tab:ablation_head_loss_training}.
We specifically investigate the impact of \emph{BNNeck}, \emph{Label Smoothing}, and \emph{Cosine Annealing}.
These techniques, inspired by their effectiveness in person re-identification, also bring consistent improvements to animal re-identification.

\begin{table}[!tbp]
\setlength\tabcolsep{4pt} 
	\caption{
	Ablation study on \emph{Head}, \emph{Loss} and \emph{Training}.
	}
	\label{tab:ablation_head_loss_training}
\vspace{-3ex}
	\begin{center}
		\resizebox{0.90\linewidth}{!}{%
			\begin{tabular}{c|cc|cc}
    		\toprule
    		\multirow{2}{*}{Method} & \multicolumn{2}{c|}{HyenaID} & \multicolumn{2}{c}{WhaleSharkID} \\
    		& R1 & mAP & R1 & mAP \\
    		\midrule
    		ARBase (\emph{w/o BNNeck})  & 64.55 & 39.23 & 44.42 & 22.29 \\
    		\midrule
    		ARBase (\emph{w/o Label Smoothing}) & 68.18 & 44.72 & 61.42 & 27.88 \\
    		\midrule
    		ARBase (\emph{w/o Cosine Annealing}) & 71.82 & 43.40 & \bftab{62.44} & 28.69 \\
            \midrule
    		ARBase(\bftab{Ours}) & \bftab{73.18} & \bftab{44.87} & \bftab{62.44} & \bftab{29.45} \\
    		\bottomrule	
			\end{tabular}
		}
	\end{center}
\vspace{-4ex}
\end{table}


\section{Conclusion}
\label{sec:conclusion}

In this study, we have tackled the complexities and challenges of animal re-identification, a task that, while conceptually similar to person re-identification, presents unique difficulties due to the diversity of species and environmental conditions.
To support research in this area, we introduced OpenAnimals, a flexible and extensible codebase tailored specifically for animal re-identification.
Through OpenAnimals, we revisited several state-of-the-art person re-identification methods and evaluated their applicability to animal re-identification benchmarks.
Our findings revealed significant gaps in the generalization of these methods, which led to the development of ARBase—a robust base model designed specifically for animal re-identification.
ARBase demonstrated superior performance across multiple benchmarks, validating the effectiveness of our animal-oriented design choices.
%
%
%
We believe that OpenAnimals and ARBase will provide a solid foundation for future advancements in animal re-identification.

{
    \small
    \bibliographystyle{ieeenat_fullname}
    \bibliography{6_reference}
}

\clearpage
\setcounter{page}{1}
\maketitlesupplementary

\section{Appendix}
\label{sec:appendix}

\begin{figure}[t]
	\centering
	\includegraphics[width=0.99\linewidth]{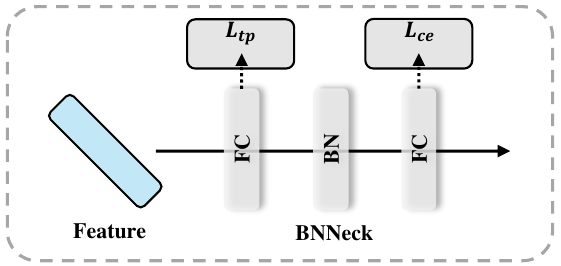}
\vspace{-1ex}
	\caption{
    Illustration of BNNeck in BoT~\cite{luo2019bag}.
    \emph{FC} and \emph{BN} for \emph{Fully-Connected Layer} and \emph{Batch Normalization},
    $L_{tp}$ and $L_{ce}$ for \emph{Triplet} and \emph{Cross-Entropy Loss}.
	}
	\label{fig:BNNeck}
\vspace{-4ex}
\end{figure}

\begin{figure}[t]
	\centering
	\includegraphics[width=0.99\linewidth]{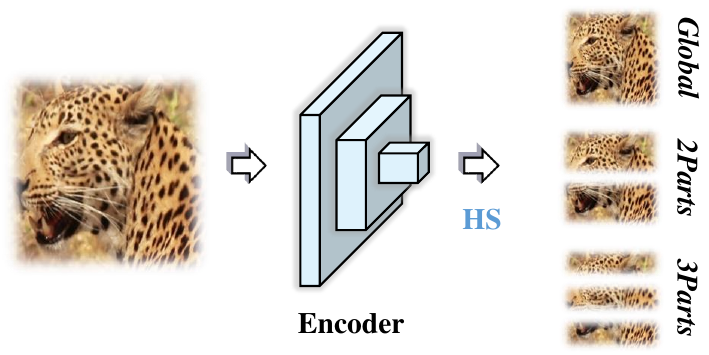}
\vspace{-1ex}
	\caption{
    Illustration of Multi-Branch Architectures in MGN~\cite{wang2018learning}.
    \emph{HS} for \emph{Horizontal Split}.
	}
	\label{fig:MGN}
\vspace{-4ex}
\end{figure}

\subsection{Illustration of BNNeck and Multi-Branch}
\emph{BNNeck}~\cite{luo2019bag} and \emph{Multi-Branch Architectures}~\cite{wang2018learning} are two fundamental modules that have demonstrated significant effectiveness in person re-identification and have been incorporated into ARBase for animal re-identification.
More specifically, BNNeck introduces a batch normalization layer that decouples the features generated by the backbone into two distinct spaces, which are then utilized for the independent computation of triplet loss and cross-entropy loss, respectively.
On the other hand, Multi-Branch Architectures~\cite{wang2018learning} employ a horizontal partitioning approach to divide the extracted features, producing fine-grained, part-level feature representations.

In the main paper, we provide a concise textual description of these modules. For further clarity, the structures and functionalities of these modules are visually illustrated in Figures~\ref{fig:BNNeck} and \ref{fig:MGN}, respectively, to assist the readers in better understanding their operations.

\begin{table}[!tbp]
\setlength\tabcolsep{6pt} 
	\caption{
	Dataset statistics. \emph{\#ID} and \emph{\#Img} for number of identities and images in each subset.
	}
	\label{tab:datasets}
\vspace{-3ex}
	\begin{center}
		\resizebox{0.90\linewidth}{!}{%
			\begin{tabular}{c|cc|cc|cc}
			\hline
			\multirow{2}{*}{Datasets} & \multicolumn{2}{c|}{Train Set} & \multicolumn{2}{c|}{Probe Set} & \multicolumn{2}{c}{Gallery Set} \\
			& \#ID & \#Img  & \#ID & \#Img & \#ID & \#Img \\
			\midrule
			HyenaID~\cite{HyenaID} 			& 145 & 1535 & 110 & 220 & 111 & 1374 \\
			LeopardID~\cite{LeopardID} 	      & 260 & 3058 & 122 & 244 & 170 & 3504 \\
			SeaTurtleID~\cite{SeaTurtuleID} 	& 224 & 3790 & 172 & 344 & 176 & 3448 \\
			WhaleSharkID~\cite{WhaleSharkID} 		& 320 & 3847 & 197 & 394 & 223 & 3452 \\
			\bottomrule
			\end{tabular}
		}
	\end{center}
\vspace{-4ex}
\end{table}

\subsection{Setup}
\subsubsection{Datasets}
In Section~\ref{sec:revisiting}, we provided a brief introduction to the datasets used in our experiments, namely HyenaID~\cite{HyenaID}, LeopardID~\cite{LeopardID}, SeaTurtleID~\cite{SeaTurtuleID}, and WhaleSharkID~\cite{WhaleSharkID}.
The detailed statistics for these datasets are presented in Table~\ref{tab:datasets}.
It is important to note that the identities used for training and testing do not overlap.

\subsubsection{Implementation Details}
%
For person re-identification methods, we adhere to their original configurations.
For ARBase, the batch size is set to $[4, 16]$ (4 identities and 16 samples per identity).
We use $m=0.3$ in Eq~\eqref{eq_triplet_ce} for the triplet loss and $\epsilon=0.1$ in Eq~\eqref{eq_label_smooth} for the cross-entropy loss.
The initial learning rate is set to $0.00035$, and the training lasts for 120 epochs.
\emph{The source code will be made publicly available.}

\subsection{More Ablation Studies}
In Table~\ref{tab:ablation_data_backbone_supp} and Table~\ref{tab:ablation_head_loss_training_supp}, we present the ablation studies of ARBase (\ie, \emph{Data}, \emph{Backbone}, \emph{Head}, \emph{Loss}, and \emph{Training \& Testing}) on two additional benchmarks, namely LeopardID~\cite{LeopardID} and SeaTurtleID~\cite{SeaTurtuleID}.
The experimental settings remain consistent with those used in the main paper.

It is important to highlight that our primary objective is to develop a robust base model that generalizes well across various species, rather than optimizing for the highest performance on a single benchmark. For instance, the highest performance on SeaTurtleID is achieved by ARBase without the use of cosine annealing.
To this end, \emph{we adopt the designs for ARBase that have demonstrated effectiveness across at least three different benchmarks.} Despite this focus on generalization, ARBase achieves state-of-the-art performance on all benchmarks, significantly outperforming the baseline models. Notably, ARBase improves the mAP by 9.32\% on SeaTurtleID compared to the best-performing baseline, as shown in Table~\ref{tab:performance_comparison}.

\begin{table}[!tbp]
\setlength\tabcolsep{4pt} 
	\caption{
	Ablation study on \emph{Data} and \emph{Backbone}.
    \emph{IBN} for \emph{Instance-Batch Normalization}, \emph{MB} for \emph{Multi-Branch Architecture}.
	}
	\label{tab:ablation_data_backbone_supp}
\vspace{-3ex}
	\begin{center}
		\resizebox{0.90\linewidth}{!}{%
			\begin{tabular}{c|c|cc|cc}
    		\toprule
    		\multirow{2}{*}{Method} & \multirow{2}{*}{\tabincell{c}{Input\\Resolution}} & \multicolumn{2}{c|}{LeopardID} & \multicolumn{2}{c}{SeaTurtleID} \\
    		& & R1 & mAP & R1 & mAP \\
    		\midrule
    		BoT~\cite{luo2019bag} 		 &  [256,128] & 54.92 & 27.65 & 84.01 & 41.92 \\
    		AGW~\cite{ye2021deep} 		 &  [256,128] & 54.10 & 28.67 & 85.17 & 46.18 \\		
    		SBS~\cite{he2023fastreid} 	 &  [384,128] & 51.23 & 26.54 & 84.01 & 44.63 \\
    		MGN~\cite{wang2018learning}  &  [384,128] & 53.69 & 28.21 & 86.05 & 46.67 \\
            \midrule
    		BoT~\cite{luo2019bag} 		 &  [384,384] & 62.70 & 34.64 & 86.05 & 49.18 \\
    		AGW~\cite{ye2021deep} 		 &  [384,384] & 60.66 & 34.55 & 88.08 & 53.80 \\		
    		SBS~\cite{he2023fastreid} 	 &  [384,384] & 59.43 & 33.12 & 86.92 & 52.40 \\
    		MGN~\cite{wang2018learning}  &  [384,384] & 61.48 & 33.64 & 88.66 & 53.52 \\
            \midrule
            ARBase(\emph{w/o IBN})       & [384,384] & \bftab{64.34} & 36.82 & 88.37 & 54.57 \\
            ARBase(\emph{w/o MB})        & [384,384] & 63.11 & 34.99 & \bftab{88.95} & 55.39 \\
    		ARBase(\bftab{Ours})         & [384,384] & \bftab{64.34} & \bftab{37.08} & 86.92 & \bftab{55.99} \\
    		\bottomrule	
			\end{tabular}
		}
	\end{center}
\vspace{-2ex}
\end{table}

\begin{table}[!tbp]
\setlength\tabcolsep{4pt} 
	\caption{
	Ablation study on \emph{Head}, \emph{Loss} and \emph{Training}.
	}
	\label{tab:ablation_head_loss_training_supp}
\vspace{-3ex}
	\begin{center}
		\resizebox{0.90\linewidth}{!}{%
			\begin{tabular}{c|cc|cc}
    		\toprule
    		\multirow{2}{*}{Method} & \multicolumn{2}{c|}{LeopardID} & \multicolumn{2}{c}{SeaTurtleID} \\
    		& R1 & mAP & R1 & mAP \\
    		\midrule
    		ARBase (\emph{w/o BNNeck})  & 56.97 & 31.43 & 76.45 & 43.44 \\
    		\midrule
    		ARBase (\emph{w/o Label Smoothing}) & \bftab{64.34} & 37.28 & 86.92 & 52.29 \\
    		\midrule
    		ARBase (\emph{w/o Cosine Annealing}) & 63.93 & \bftab{37.78} & \bftab{89.53} & \bftab{56.52} \\
            \midrule
    		ARBase(\bftab{Ours}) & \bftab{64.34} & 37.08 & 86.92 & 55.99 \\
    		\bottomrule	
			\end{tabular}
		}
	\end{center}
\vspace{-4ex}
\end{table}

\subsection{Visualization}
\begin{figure}[t]
	\centering
	\includegraphics[width=0.99\linewidth]{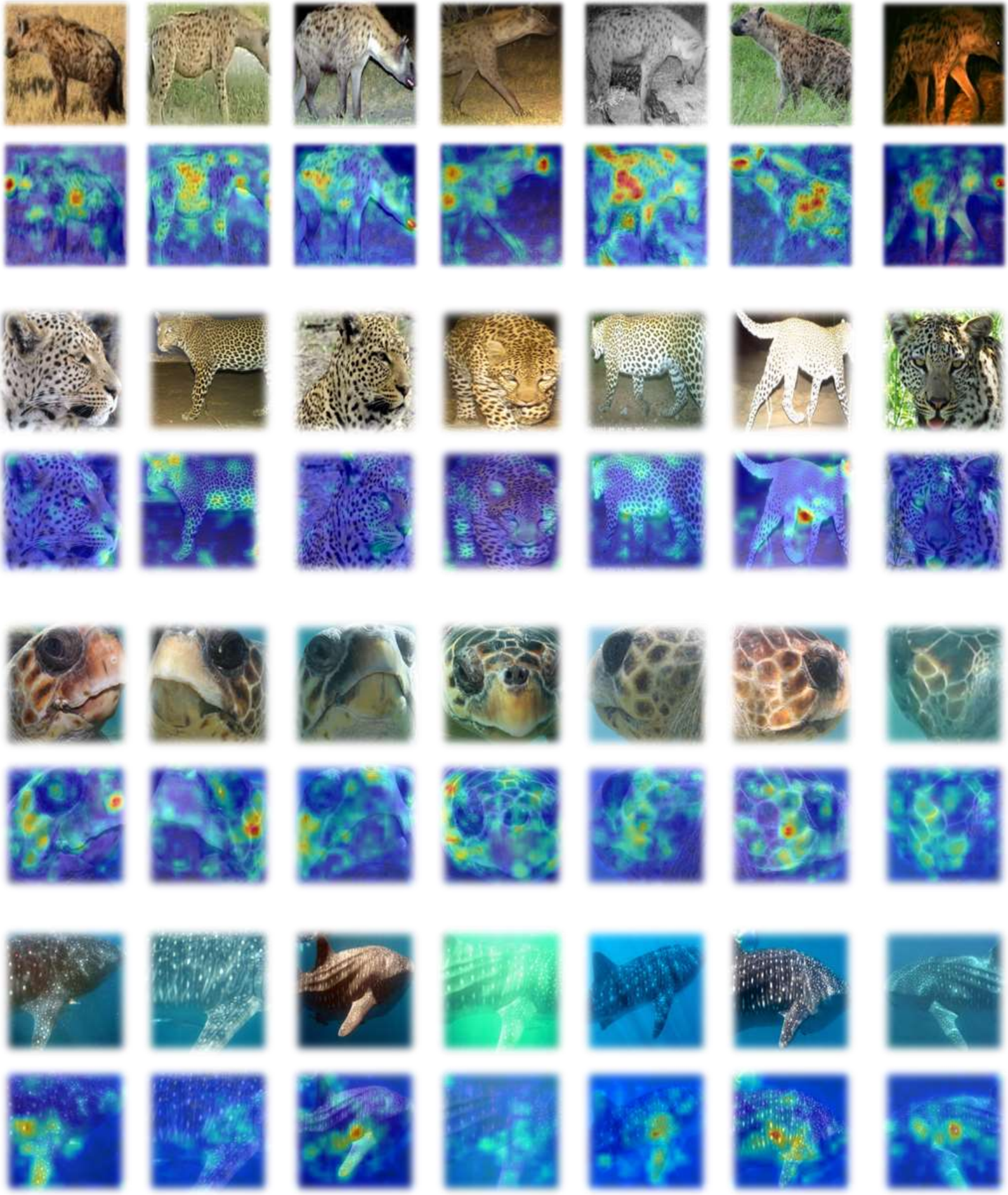}
\vspace{-1ex}
	\caption{
    Visualization of heatmaps.
    These heatmaps are obtained by applying max pooling along the channel dimension of the features extracted by the backbone.
	}
	\label{fig:visualization}
\vspace{-2ex}
\end{figure}

In Figure~\ref{fig:visualization}, we present the heatmaps generated by the backbone of ARBase trained on each benchmark, highlighting the key regions used for the re-identification of each species.
From these visualizations, it is evident that animal re-identification predominantly relies on body or head texture features. This observation is intuitive, as individual animals exhibit unique texture patterns, much like human fingerprints.

Interestingly, this finding is consistent with conclusions from biological research~\cite{hiby2009tiger}, further validating the critical role of texture features in distinguishing individual animals across species.

\subsection{Discussion and Future Work}
Our study makes a meaningful contribution towards advancing animal re-identification, yet this task warrants further exploration.
Here, we identify some promising directions for future research:
\begin{enumerate}[label=(\alph*)]
    \item \textbf{Attribute-assisted Animal Re-Identification}: Semantic attributes, such as gender and age, are useful auxiliary tools for person re-identification. For animal re-identification, summarizing long-term attributes could enhance identity recognition.
    \item \textbf{Video-based Animal Re-Identification}: Due to the challenges associated with data collection and annotation, current benchmarks for animal re-identification are primarily image-based. Videos, however, provide richer information and could be more promising for accurate animal re-identification.
    \item \textbf{Generalizable Animal Re-Identification}: Animal re-identification involves various species, making it valuable to develop a generalized model capable of recognizing multiple species. This is particularly feasible with the emergence of Large Language Models (LLMs) that encapsulate rich knowledge across different species.
\end{enumerate}

\end{document}